\documentclass[]{OAGM}
\OAGMarXiv{1404.3538}

\usepackage{graphics}
\usepackage{authblk}

\title{Improving weather radar by fusion and classification}

\author[1]{H.~Ganster}
\author[1]{M.~Uray}
\author[1]{S.~Steginska}
\author[2]{G.~Croonen}
\author[3]{R.~Kaltenb{\"o}ck}
\author[4]{K.~Hennermann}
\affil[1]{JOANNEUM RESEARCH Forschungsgesellschaft mbH, Austria}
\affil[2]{AIT Austrian Institute of Technology GmbH, Austria}
\affil[3]{Austro Control, {\"O}sterreichische Gesellschaft f{\"u}r Zivilluftfahrt mbH, Austria}
\affil[4]{MeteoServe Wetterdienst GmbH, Austria}

\begin{document}
\maketitle

\begin{abstract}
In air traffic management (ATM) all necessary operations (tactical planing, sector
configuration, required staffing, runway configuration, routing of approaching aircrafts) rely
on accurate measurements and predictions of the current weather situation. An essential basis
of information is delivered by weather radar images (WXR), which, unfortunately, exhibit a vast
amount of disturbances. Thus, the improvement of these datasets is the key factor for more accurate
predictions of weather phenomena and weather conditions. Image processing methods based on texture
analysis and geometric operators allow to identify regions including artefacts as well as zones
of missing information. 
Correction of these zones is implemented by exploiting multi-spectral satellite
data (Meteosat Second Generation). Results prove that the proposed system for artefact detection
and data correction significantly improves the quality of WXR data and, thus, enables more reliable
weather now- and forecast leading to increased ATM safety.
\end{abstract}

\section{Motivation}
\begin{figure}[ht]
  \centering
		{ \resizebox{0.43\textwidth}{0.32\textwidth}{\includegraphics{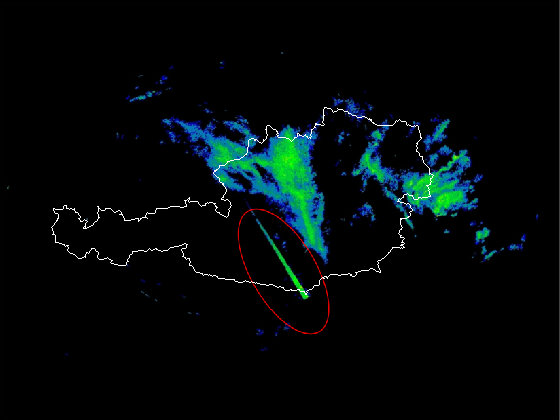}} } \hspace{5mm}
    { \resizebox{0.43\textwidth}{0.32\textwidth}{\includegraphics{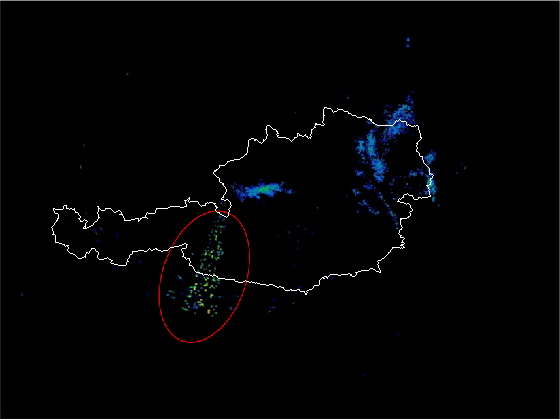}} } \\
  \caption{Weather radar images affected by typical disturbances: RLAN artefact (left) and disturbance due to in-flight radar (right).}
  \label{fig:wxrartefacts}
\end{figure}

Air traffic controllers face the challenge to cope with a huge amount of information. Significant sources
are found in images and other sensors measuring the current weather conditions. By using weather
radar (WXR), one can predict three dimensional evaluation and forecast of precipitation, danger of
icing, storm expansion and development as well as its strength {\cite{Kaltenbock2005}}. But due to
the mountainous region of Austria, the limited number of stations, obstructions arising in close
range to the transmitter, a limited sampling possibility, beam spread, and other signal-absorbing effects
it is not possible to cover the complete Austrian area and/or the images are afflicted by
interferences. Fig.~\ref{fig:wxrartefacts} gives two examples of such artefact appearance in WXR
imagery, displayed as composite of maximum projected weather radar reflectivity derived from $4$ radar 
sites for the Austrian airspace.
The classes of artefacts include anomalies like clutter (sea and ground clutter), false echoes (caused
by aircrafts, birds or insects, dust particles, atmospheric turbulence, the sun or other radars), 
bright bands (caused by height reflectivity of a cloud's melting zone) and beam blockage (shadowing, concealed zones).

\begin{figure}[ht]
  \centering
    \resizebox{0.9\textwidth}{0.4\textwidth}{\includegraphics{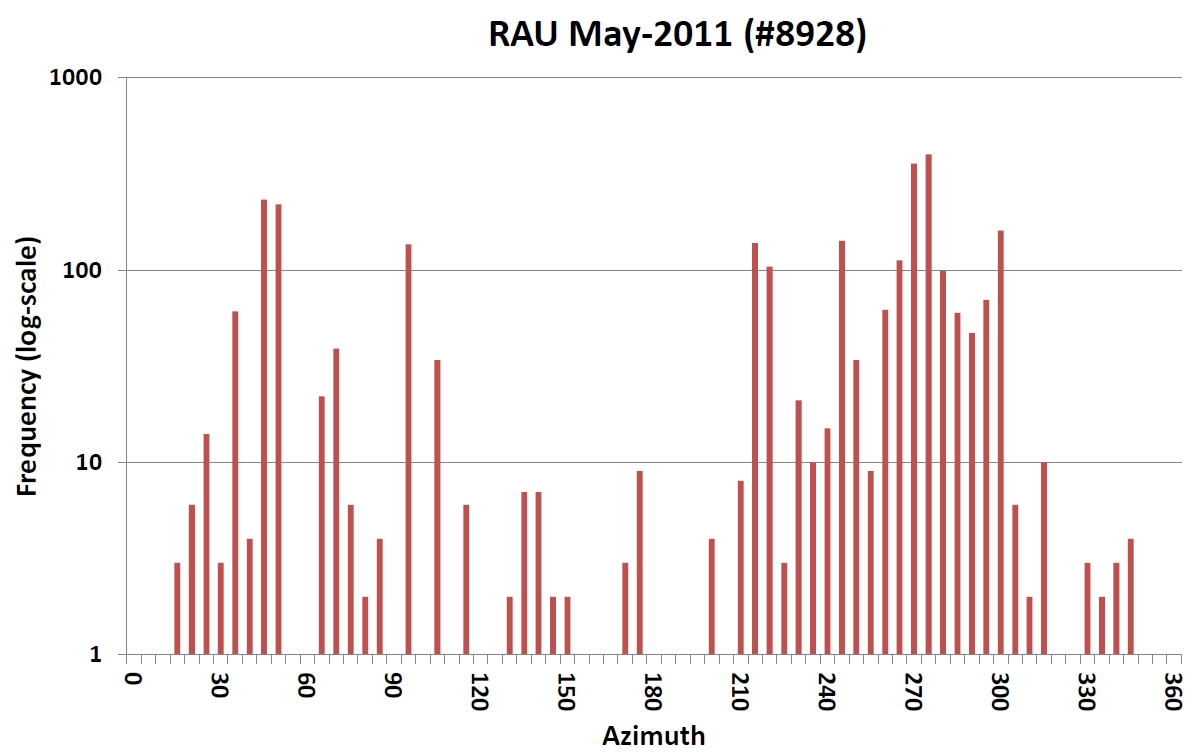}} \\
  \caption{Frequency of RLAN disturbances per direction for radar station Rauchenwarth in Mai 2011.}
  \label{fig:rlanfreq}
\end{figure}

A statistic evaluation\footnote{Data provided by Austro Control.} of WXR images from one single radar station during a period of one month (May 2011) showed that artefacts due to 'Radio Local Area Network' (RLAN) disturbances were found in $2713$ from $8928$ images (i.e.~$30.4\%$ afflicted images, Fig.~\ref{fig:rlanfreq}). Due to the fact that RLAN disturbances are an international problem, European-wide activities to eliminate this type of artefacts are already pursued (e.g.~\cite{Horvath2008}).
This supports the fact that the improvement of WXR images - by combining them with other sensors in
order to complete concealed zones and eliminate interfering signals without losing important
meteorological information - is essential for an accurate prediction of weather phenomenons and
atmospheric conditions (see also~\cite{Verworn2008}), which is furthermore an essential factor 
in the work process of air traffic controllers.

In recent literature some efforts to enhance the quality of radar data by detecting and removing these unwanted
disturbances can be found: Gill~\cite{Gill2008} presents a method for clutter removal. The method
'Elevation angle Dependent Co-Occurrence Matrix EDCM' exploits the property, that clutter, due to its surface-induced origin, usually is associated with the radar's lower elevations.
Ziyang et al.~\cite{Ziyang2008} propose a noise filtering algorithm based on fuzzy features and Support
Vector Machines (SVM), which effectively allows the removal of ground clutter or clear sky echoes.
Peura~\cite{Peura2002} develops a dedicated set of detectors according to the respective artefact
class. Tools for the detection of birds and insects, speckle noise, line segments (external emitter and
sun) or ships and airplanes are given. Lakshmanan and Zhang~\cite{Lakshmanan2009} focus on
artefacts caused by biological contaminants: insects, bats and birds employing clustering,
segmentation, and neural networks (NN).

In order to correct disturbed areas, data from other sources has to be included. Most notable is
weather satellite data, where first activities on 'gap filling' are reported (\cite{Bovith2006},
\cite{Michelson2004}). Satellite data for Europe is available in 12 spectral channels and a temporal 
resolution of 15 minutes.

The work presented here does not focus on specific classes of artefacts, but takes a more generalized approach
by targeting several classes by a fusion-based concept of artefact detection and correction. In
this concept texture-based processing is combined with geometry-related detectors, which
furthermore are subject to a spatio-temporal analysis (Section~\ref{sec:WXR_artefact}). 
The affected areas as well as concealed zones in WXR data are completed in a meteorological
reasonable manner with the help of satellite data employing classification approaches
(Section~\ref{sec:WXR_correction}).

\section{Artefact Detection}\label{sec:WXR_artefact}
WXR images represent precipitation values as $4$ bit data using $14$ reflectivity quantification values\footnote{Reflectivity values: $0$, $11.82$, $14$, $19.46$, $22$, $26.69$, $30$, $34.19$, $38$, $41.82$, $46$, $50.19$, $54.27$ and $58 dBZ$}, where higher labels represent more intense zones of precipitation. The resolution of WXR images is $1 km \times 1 km$ resulting in images of the size $824 \times 648$ for Austria.

Since artefacts in WXR images arise in a variety of appearances, their detection is based on a
concept of fusing texture-based methods with geometry-related analysis. Texture-based detection is
able to deliver a segmentation of the more general appearances of artefacts, whereas the
geometry-related detection utilizes the circular properties of radar imaging.

\subsection{Texture segmentation}
\begin{figure}[ht]
  \centering
		{ \resizebox{0.35\textwidth}{0.35\textwidth}{\includegraphics{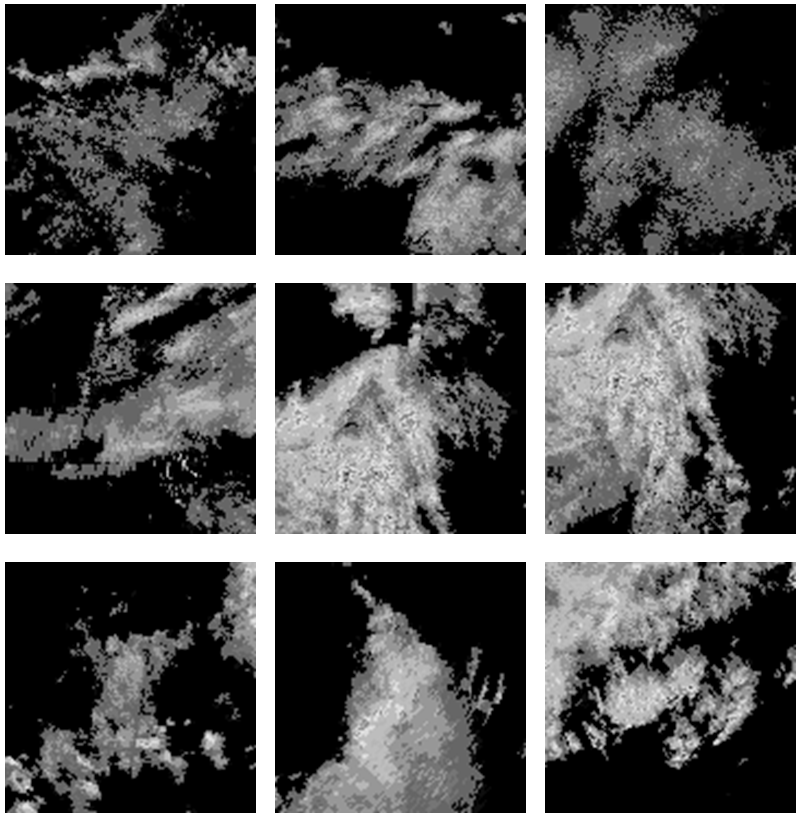}} } \hspace{10mm}
    { \resizebox{0.35\textwidth}{0.35\textwidth}{\includegraphics{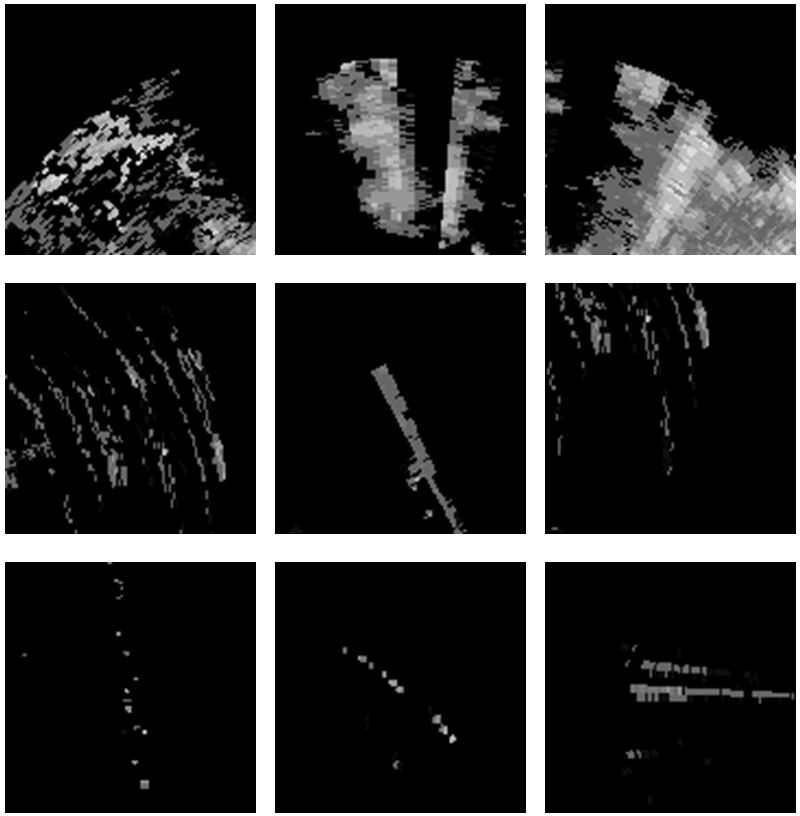}} } \\
  \caption{Texture patches of precipitation (left) and disturbances (right) used in training for texture segmentation.}
  \label{fig:texture}
\end{figure}

Based on the available WXR images\footnote{Data from one year, with a capturing interval of 5
minutes.}, 460 texture patches of zones with artefacts as well as normal precipitation were
extracted serving as data for the selection of the best texture segmentation algorithm
(Fig.~\ref{fig:texture}). 
The variety of analysed texture segmentation methods included variants
ranging from 'Grey Level Co-Occurrence Matrix' (\cite{Haralick1973}) and 'Region Covariance
Descriptor' (\cite{Porikli2005}) over 'Gabor Filters' in combination with other descriptors
(\cite{Tou2009}) to 'Leung-Malik Set' (\cite{Leung2001}) and 'Maximum Response Set'
(\cite{Varma2004}). Best results were achieved with a covariance descriptor on feature images of Gabor filter outputs (\cite{Steginska2012}) by using six orientations and three frequencies resulting in
18 filter outputs on a neighbourhood of 39 pixels. The evaluation was performed with 'Nearest
Neighbour Classification' and 'Leave One Out Cross Validation'.

\subsection{Geometric analysis}
\begin{figure}[ht]
  \centering
    { \resizebox{0.43\textwidth}{0.32\textwidth}{\includegraphics{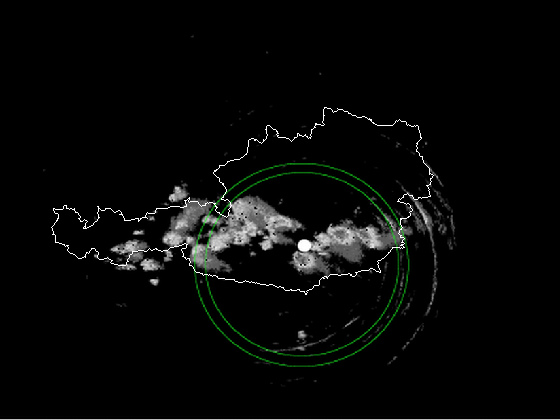}} } \hspace{5mm}
    { \resizebox{0.43\textwidth}{0.32\textwidth}{\includegraphics{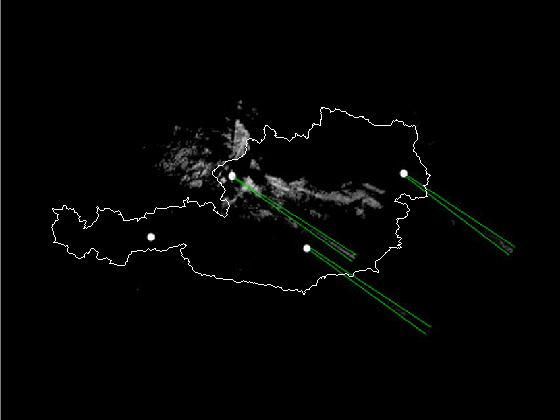}} } \\
  \caption{Illustration of geometric properties of artefacts.}
  \label{fig:geometry}
\end{figure}

A certain amount of artefacts exhibit specific geometric appearances due to the principles of radar
imaging and appear either as linear structures pointing towards the radar station or circular
structures with the radar station as the centre (Fig.~\ref{fig:geometry}). The detection of these specific 
artefact classes is achieved with a combination of Hough-transformation and mathematical morphology 
utilizing properties in polar coordinates.
The complete sector of linear structures (Fig.~\ref{fig:geometry} right) transforms to a vertical line in polar coordinates. By morphological closing and filtering for elongated vertical structures these artefacts are extracted in polar space. A reprojection to cartesian coordinates delivers the required structures.
The circular structures tend to be composed by a number of small objects instead of connected circular regions. Thus, application of circular Hough transformation is best suited to detect these accumulated circular structures.
Since the Hough space can be constrained to the parameters of the radar stations ($4$ fixed centres and operating distance of $225km$) and does not need a general search of the parameter space the application of Hough transformation stays computationally reasonable.

\subsection{Fusion and performance}
For evaluation of the artefact detection the necessary ground truth was established by manual pixel-based
segmentation of 107 complete WXR images. Segmentation was done independently by three
meteorologists and majority voting per pixel. 

Texture segmentation alone could achieve a performance of $83,5\%$\footnote{Note that all performance values given do not consider
the background class (no precipitation or artefact in the image) as this would lead to tremendously
higher numbers, since background pixels (i.e. no precipitation) are more common than foreground pixels.} correctly classified pixels with
a sensitivity of $96.0\%$ and specificity of $85.8\%$. 
The results indicate an already correct segmentation of artefact pixels, but still have a high
number of false positives, which in turn would lead to too much filtering of precipitation in the
following step of image correction. Thus, the texture-based segmentation was fused with the
geometry-based detection and furthermore subjected to a spatio-temporal analysis.

Since artefacts tend to appear only for short periods of time in the same spot, the analysis of
predecessor images significantly reduced the false positive detection. The inclusion of only two
more timestamps in the fused system led to a tremendous improvement in performance with a number of
$97.8\%$ correctly classified pixels for image sets, where all predecessor images were available
with the standard time frame of 5 minutes. In some cases, due to hardware issues, WXR images are
missing. Consequently, the predecessor images exhibit a longer timespan (e.g. $10$ minutes). But also in
these cases the fused system delivered an improvement in comparison to pure texture-based
segmentation with $94.8\%$ correctly classified pixels. 

As drawback of the fused system, the number of false negatives, i.e. artefact pixels that could not
be detected, is higher than in the pure texture-based system leading to a sensitivity of $84.8\%$ and 
a specificity of $99.8\%$. Considering the application domain of air traffic
management, it is essential that no significant precipitation is lost in the process. Thus, the
trade-off in artefact detection is justified and the overall achieved system performance is
$96.03\%$.

\section{Weather Radar Correction}\label{sec:WXR_correction} 
\begin{figure}[ht]
  \centering
    { \resizebox{0.43\textwidth}{0.32\textwidth}{\includegraphics{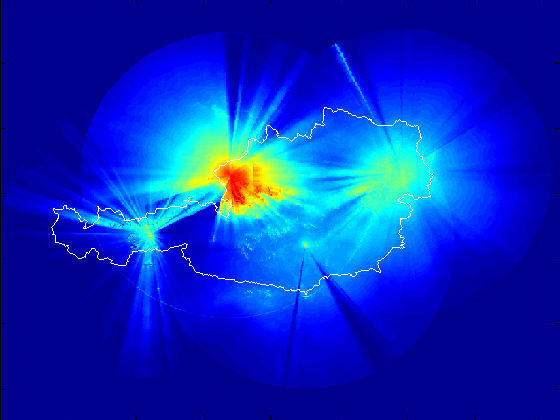}} } \hspace{5mm}
    { \resizebox{0.43\textwidth}{0.32\textwidth}{\includegraphics{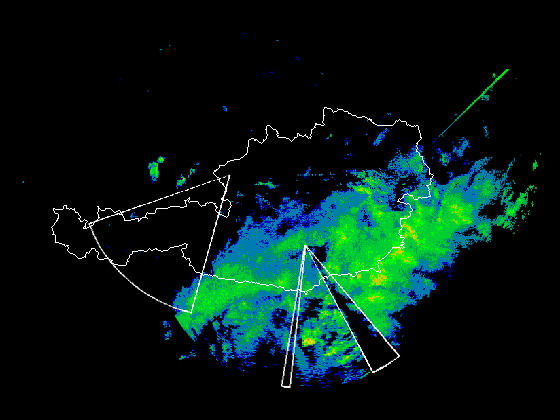}} } \\
  \caption{Sum of precipitation in weather radar images for one year (left) and derived shadow sectors overlaid on WXR image (right).}
  \label{fig:wxrsum}
\end{figure}

Besides the removal of spurious zones in weather radar images, there are also concealed zones,
mostly due to orographic shadowing. These zones appear as circle segments and are easily detected by
an analysis of the precipitation sum over a year (Fig.~\ref{fig:wxrsum} left).
The shadow mask for each WXR image is derived by detection of linear boundaries of the current precipitation and verifying if this linear boundary corresponds to a boundary from the previously detected shadow zones in the summation image (Fig.~\ref{fig:wxrsum} right). The mask derived from this analysis in combination with the mask derived from the artefact
detection (Section~\ref{sec:WXR_artefact}) represents the overall correction mask of non-valid
WXR pixels and builds the basis for the desired correction. The correction task itself is performed by utilizing
the channels of the 'Meteosat Second Generation (MSG)' satellite. The underlying hypothesis for the correction is
that similarities in MSG images point to similar WXR values. Nevertheless, the locality of weather phenomena and orographic influences do not allow for a global classification of these regions. Local analysis is necessary in order to fill
all non-valid pixels with meteorologically reasonable values.

In order to obtain local training data, each connected region in the correction mask is extracted by connected component labelling. The training set around each connected region is then obtained by repeated application of the morphological operation of dilation followed by subtraction of the original region mask. Consequently, training is performed for each region separately with the pixels obtained from its outside border.

\subsection{Classification}
Several classification concepts were evaluated for filling the concealed regions regarding the flexibility of the local approach. The drawback of the local approach is that less data is available for training, which made the application of more advanced classifiers such as SVM less successful. The k-nearest neighbour classifier, although being a simple algorithm, provided the best correction values at low computation times required for the online application. The classifier employs the $12$ MSG channels for each pixel inside the region to be corrected and searches the training set for pixels with similar MSG-vectors. After calculating the Euclidean distance between the pixel to be corrected and the MSG-vectors available for training,
the top-$k$ neighbours of minimum distance return $k$ WXR-labels. From these $k$ WXR-labels the
mean value is used to correct the desired pixel.

\subsection{Evaluation of correction results}
\begin{figure}[ht]
  \centering
    { \resizebox{0.84\textwidth}{0.28\textwidth}{\includegraphics{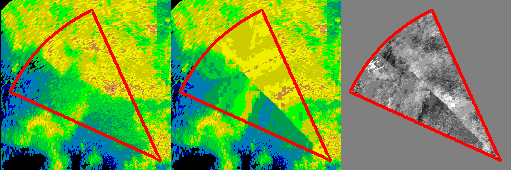}} } \\
  \caption{Results of correction by classification in the marked regions. From left to right: original ground truth, correction result and error image.}
  \label{fig:mask_and_correction}
\end{figure}

Evaluation of correction performance was done for simulated regions of three different sizes
(small, medium and large), respectively containing 100, 1000 and 10000 pixels. 
Positioning was done on valid regions inside the original WXR images, so that the removed data 
could directly be used as ground truth data for evaluation of correction quality.
Allowing for plus or minus one intensity class error, WXR correction was correctly performed for
$80\%$ of the pixels for small, $86\%$ of the pixels for medium and $72\%$ of the pixels for large
regions. As an example a snapshot of the confusion matrix for the medium sized evaluation is given
in Table~\ref{tab:conf_med}\footnote{High labels from $8$ to $12$ were left out for illustration clarity as all values were $0$ reflecting that severe precipitation is quite rare and those values did not appear in the set of test data.}.
\begin{table}[!h]
\tiny
\caption{Excerpt of confusion matrix of corrected weather radar value for medium sized regions.}
\label{tab:conf_med}
\begin{center}
\begin{tabular}{r|cccccccccccccccc}
\hline\hline 
- & 0&1&2&3&4&5&6&7 \\
\hline
0 & 0.93 & 0.31 & 0.11 & 0.02 & 0.02 & 0.01 & 0.01 & 0.00 \\
1 & 0.05 & 0.34 & 0.15 & 0.03 & 0.02 & 0.00 & 0.00 & 0.00 \\
2 & 0.02 & 0.29 & 0.52 & 0.24 & 0.10 & 0.03 & 0.02 & 0.00 \\
3 & 0.00 & 0.03 & 0.15 & 0.36 & 0.20 & 0.07 & 0.03 & 0.01 \\
4 & 0.00 & 0.02 & 0.06 & 0.27 & 0.42 & 0.27 & 0.13 & 0.06 \\
5 & 0.00 & 0.01 & 0.01 & 0.06 & 0.18 & 0.34 & 0.26 & 0.09 \\
6 & 0.00 & 0.00 & 0.00 & 0.01 & 0.05 & 0.23 & 0.40 & 0.50 \\
7 & 0.00 & 0.00 & 0.00 & 0.00 & 0.01 & 0.04 & 0.13 & 0.31 \\
\hline \\[-0.8em]
GT+/-1 & 0.98 & 0.95 & 0.82 & 0.87 & 0.80 & 0.85 & 0.80 & 0.84 \\
\hline \\[-0.8em]
Sample size  & 71343 & 4653 & 15501 & 4592 & 4040 & 1179 & 875 & 176 \\
\hline
\end{tabular}
\end{center}
\end{table}
From the results it is apparent that WXR correction using a k-nearest neighbour classifier works
best for small to medium sized regions supporting the hypothesis of locality in the correction
process. An exception are zones with higher precipitation rates, as they are generally smaller and
less often present in the training sets. These effects lead to an underestimation of the
precipitation rates in the corrected regions. Another important aspect of correction is the
meteorological plausibility of the appearance of the correction. To illustrate the quality of the results 
Fig.~\ref{fig:mask_and_correction} displays a sample region used during classification evaluation.
Dark areas in the error images denote negative errors, whereas light areas denote positive errors. In the displayed large mask a maximum error of $4$ is achieved. For medium sized masks the error does not exceed $0.9$.

For further improvement of the correction results it is planned to include information from tracking precipitation cells over time instances into concealed regions as well as other sources such as numerical weather prediction (NWP) data (e.g.~temperature, pressure fields etc.).

\section{Conclusion}
\begin{figure}[ht]
  \centering
    { \resizebox{0.43\textwidth}{0.32\textwidth}{\includegraphics{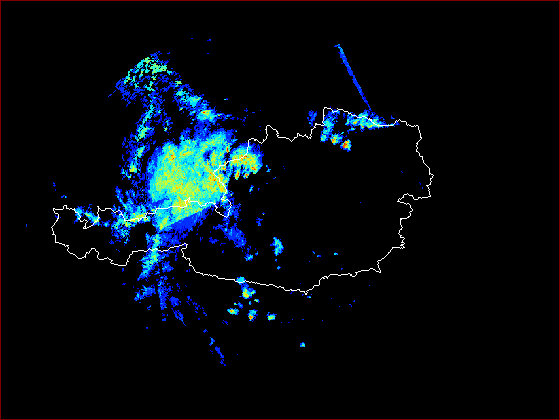}} } \hspace{5mm}
    { \resizebox{0.43\textwidth}{0.32\textwidth}{\includegraphics{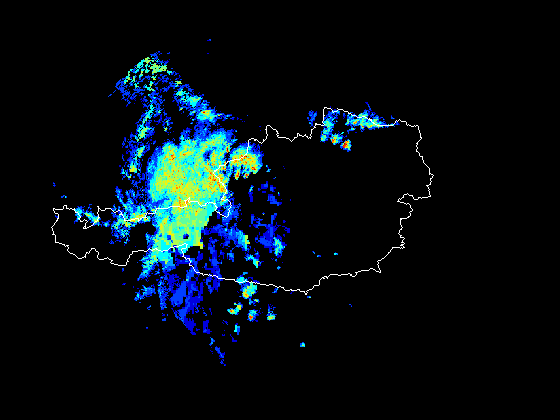}} } \\
  \caption{Original weather radar image (left) and result of weather radar improvement by image processing (right). }
  \label{fig:result}
\end{figure}

This paper presents a system that is able to improve weather radar images in a meteorologically
reasonable manner consisting of two components: artefact detection and removal as well as
completing concealed zones. The system led to filing of two positively evaluated patents and is
currently under evaluation in a semi-operational environment at 'Austro Control GmbH', where the results are compared 
to the original WXR images in parallel operation.
Therefore, a web-based logbook-tool has been established that gives feedback on the quality of the corrected WXR images. Fig.~\ref{fig:result} displays the final result of the image processing chain for weather radar improvement, where areas with classification 
values are pixelated on a coarser scale for better illustration. It clearly shows the
regions of removed artefacts in the North and the completed zones in the South-West.

In the future it is planned to integrate the system directly at the sensor site (WXR station)
allowing for the corrected WXR images to be the basis for all meteorological short term forecasting.
This will ease the detection of significant meteorological indicators, like effects from wind shear
as precursors to heavy thunderstorms. The improved forecasts will in turn have significant effects
on safety in aviation.

\section*{Acknowledgments}
This research was partly funded by BMVIT under programme TAKE OFF, project nrs. $820742$, $825601$ and $838987$.

\begin{small}
\bibliography{textur_v2}
\end{small}
\end{document}